\documentclass[conference]{IEEEtran}

\IEEEoverridecommandlockouts                                                                                        

\usepackage[numbers]{natbib}
\usepackage{times}
\usepackage{siunitx}
\usepackage{gensymb}
\usepackage{lipsum}
\usepackage{subcaption}
\usepackage{multicol}
\usepackage{graphicx}
\usepackage{wrapfig}
\usepackage{float}
\usepackage{tikz}
\usepackage{mathtools}
\usepackage{booktabs}
\usepackage[bookmarks=true,hidelinks]{hyperref}
\usepackage[noabbrev,capitalize]{cleveref}
\usepackage{amsmath}
\usepackage{mleftright}
\usepackage{algorithm,algpseudocode}
\usepackage{placeins}
\usepackage{booktabs}
\usepackage{overpic}
\usepackage{float}

\definecolor{linkcolor}{HTML}{648EB0}
\newcommand{\link}[2]{\href{#1}{\textcolor{linkcolor}{#2}}}

\newlength{\figwidth}          \newlength{\figheight}          
\newcommand{\midtilde}{\raisebox{-0.25em}{\textasciitilde}}

\usepackage{soul} \usepackage{xcolor} 
\definecolor{softyellow}{RGB}{255,255,204} 
\sethlcolor{softyellow}

\usepackage[utf8]{inputenc}
\usepackage{pgfplots}
\DeclareUnicodeCharacter{2212}{−}
\usepgfplotslibrary{groupplots,dateplot}
\usetikzlibrary{shapes,arrows.meta,positioning,patterns,shapes.arrows}
\pgfplotsset{compat=newest}

\makeatletter
\renewcommand*{\NAT@spacechar}{~}
\makeatother

\begin{document}

\title{Safe \& Accurate at Speed with Tendons: \\A Robot Arm for Exploring Dynamic Motion}

\author{Simon Guist$^{1}$, Jan Schneider$^{1}$, Hao Ma$^{1}$, Le Chen$^{1}$, Vincent Berenz$^{1}$,  Julian Martus$^{2}$, Heiko Ott$^{1}$, Felix Grüninger$^{2}$,\\ Michael Muehlebach$^{1}$, Jonathan Fiene$^{2}$, Bernhard Schölkopf$^{1}$ and Dieter Büchler$^{1}$
\thanks{$^{1}$Max Planck Institute for Intelligent Systems, 72076 Tübingen, Germany.
        {\tt\small firstname.lastname@tuebingen.mpg.de}}
\thanks{$^{2}$Max Planck Institute for Intelligent Systems, 70569 Stuttgart, Germany.
        {\tt\small lastname@is.mpg.de}}
}

\makeatletter
\let\@oldmaketitle\@maketitle
\renewcommand{\@maketitle}{\@oldmaketitle
  \centering
  \includegraphics[width=0.9\textwidth]     {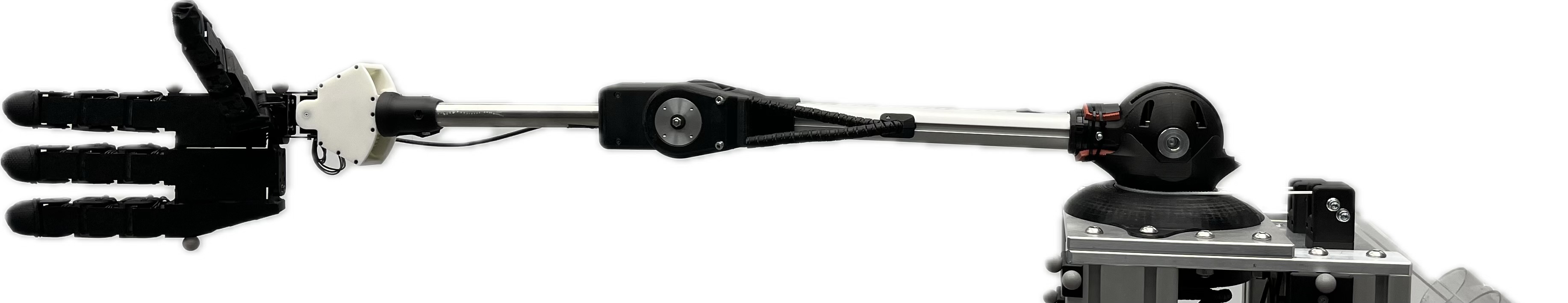}
    \vspace{-.25cm}

    }
\makeatother

\maketitle

\begin{abstract}
Operating robots precisely and at high speeds has been a long-standing goal of robotics research.
Balancing these competing demands is key to enabling the seamless collaboration of robots and humans and increasing task performance.  
However, traditional motor-driven systems often fall short in this balancing act.
Due to their rigid and often heavy design exacerbated by positioning the motors into the joints, faster motions of such robots transfer high forces at impact. 
To enable precise and safe dynamic motions, we introduce a four degree-of-freedom~(DoF) tendon-driven robot arm. 
Tendons allow placing the actuation at the base to reduce the robot's inertia, which we show significantly reduces peak collision forces compared to conventional robots with motors placed near the joints.
Pairing our robot with pneumatic muscles allows generating high forces and highly accelerated motions, while benefiting from impact resilience through passive compliance. 
Since tendons are subject to additional friction and hence prone to wear and tear, we validate the reliability of our robotic arm on various experiments, including long-term dynamic motions. 
We also demonstrate its ease of control by quantifying the nonlinearities of the system and the performance on a challenging dynamic table tennis task learned from scratch using reinforcement learning. 
We open-source the entire hardware design, which can be largely 3D printed, the control software, and a proprioceptive dataset of 25 days of diverse robot motions at \link{https://webdav.tuebingen.mpg.de/pamy2/}{webdav.tuebingen.mpg.de/pamy2}.
\end{abstract}
\IEEEpeerreviewmaketitle

\section{Introduction}
\label{sec:introduction}

Tasks such as playing table tennis, harvesting delicate berries, or carrying heavy objects differ inherently in their force, precision, and compliance demands.  
Humans naturally excel at each of these tasks by encompassing a broad array of various motion characteristics, including slow and precise motions, high-force trajectories, and fast as well as highly accelerated movements.
Replicating the full range of these capabilities, along with the inherent safety properties of human arm movements, such as compliance and backdrivability, has been a challenging endeavor in robotics research.

The desired robot capabilities can be roughly divided into (i) the achievable speed and (ii) force, (iii) the closeness to human size, (iv) the ease of control, and (v) the safety properties of the robot and environment at impact. 
High speed and force are essential to accomplish tasks quickly while potentially handling heavy objects. 
Executing accurate motions with high safety standards enables handling delicate objects or tasks with low tolerance for imprecision, such as in manufacturing.  
Inherent safety extends the set of allowed contacts and enables the control algorithm to be less conservative and take more risks to optimize for performance.
In that manner, robots could move at higher speeds due to the reduced negative consequences of unintended impacts. 
The amount of anthropomorphism is crucial when robots should act in human-created environments and tasks, potentially even working alongside humans.
The size of the robot affects all categories: while smaller robots can be precise, fast, and safe, executing high forces is rather difficult. 
On the other hand, a big robot can generate forceful and fast trajectories but sacrifices safety at impact.  

Distinct robot designs excel in each of these categories. 
Industrial motor-driven robots are traditionally heavy and rigid; these properties are necessary to describe the system precisely using the rigid body dynamics equations, which can be used to easily attain high-quality control.
Paired with strong motors, industrial robots additionally excel at maximum force and speed. 
On the downside, industrial robots easily cause damage to themselves and the environment in collisions. 
Collision avoidance for such robots often necessitates the use of 1) environmental sensors and tracking systems, 2) limiting the robot's speed, or 3) range of motion. 
Still, these techniques may not prevent all collisions while constraining the robot's performance. 
For this reason, collaborative robots or ``cobots'' have been invented. 
These robots are relatively slow and weak but safe for human interaction, compared to industrial robots that are fast and precise but prone to damage upon collision. 
Alternatively, the negative effects of collisions could be significantly reduced by using entirely soft robots. 
These systems often rely on stretchable components as well as compressible fluids.
However, continuous scratching along the surfaces of such materials can lead to damage since soft robots are generally less durable than rigid robots. 
Moreover, accurate control of fully soft robots~\cite{wehner2016integrated}, tends to be problematic.

One way to find a good tradeoff between these desired robot capabilities is to use tendon drives. 
Tendon drives allow building robot arms with minimal moving masses by transferring the actuation to the robot base. 
Such robot designs can be safe, even when operating at increased velocities, due to their inherent backdrivability and low inertia.
Paired with powerful actuation, such systems can emit high forces and reach high speeds. 
By positioning the actuation at an arbitrary location in the robot body, the system can more easily take any desired form to approach anthropomorphism.  
A major drawback of tendon-driven robots is its challenging control. 
Repeated high-force transmission wears out the tendon guidance, and high amounts of friction typically add nonlinearities and stochasticity. 

\begin{figure}
    \includegraphics[width=\columnwidth]{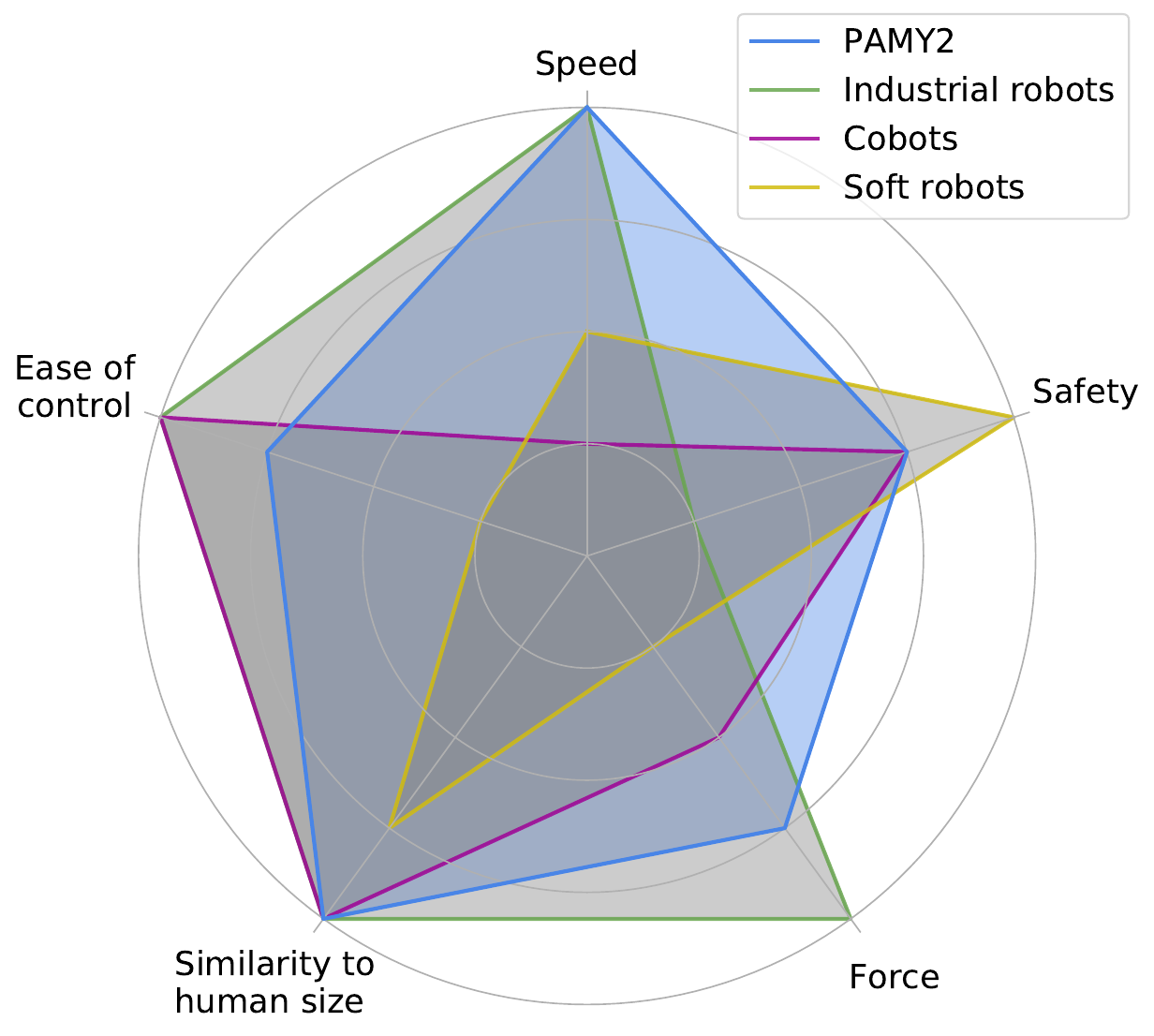}
    \caption{Visualization of the capabilities of different robot designs. Industrial robots excel in speed, generated forces, and ease of control but are not safe to operate in the proximity of humans.
    Cobots are easy to control and safe but sacrifice speed and force. Soft robots are generally superior in terms of safety but are hard to control and are often unable to generate high forces. Our robot, PAMY2, is capable of generating high-force and high-velocity trajectories while being significantly safer than most robots.
    Furthermore, the reduced friction makes our robot easier to control than typical tendon-driven systems.}
    \label{fig:robot_comparison}
\end{figure}

In this work, we present PAMY2, a durable and fast 4-DoF tendon-driven robot arm roughly of the size of a human arm.
PAMY2 features significantly lower friction than previous designs.
We pair PAMY2 with powerful pneumatic artificial muscles~(PAMs).
PAMs offer the advantage of avoiding stiff joints, which results in less severe peak forces at collisions. 
At the same time, this type of actuator can generate great forces to achieve either fast motions or lift heavy objects. 
PAMY2 incorporates new low-friction tendon guidances and ball bearings in the joints to improve the ease of control. 
To that end, we show that our system is more linear than the tendon-driven arm most similar to ours~\cite{buchler_lightweight_2016,buchler_control_2018}.
Moreover, we ensured through various hardware iterations that our design is resilient. 
We let the robot run for 25 days uninterrupted while quantifying the repeatability of the system throughout.
Additionally, we illustrate that our system produces similar impact forces to the Franka Panda and UR5e at $\midtilde\,4\times$ the speed. 
To illustrate the ease of control of PAMY2, we learn table tennis smashes with reinforcement learning~(RL) as in~\cite{buchler_learning_2022} and double the ball's speed while simultaneously improving precision.
The setting is identical down to the hyperparameters, showing that by just using our robot, performance improved substantially.
This result is especially interesting since PAMs are nonlinear actuators that change their dynamics with temperature and exhibit hysteresis effects~\cite{tondu_modelling_2012} that often require advanced modeling techniques~\cite{buchler_control_2018,shaj_action-conditional_2021,shaj2022hidden}.
\cref{fig:robot_comparison} puts our new robot, PAMY2, into perspective with the other discussed robot designs in terms of the desired robot capabilities. 

To accelerate progress on learning for dynamic tasks, we \textit{open-source the design} as well as the \textit{entire software infrastructure} required to run our system, including a C++ and Python interface. 
Since our system uses mainly off-the-shelf, commercially available components, 3D-printed parts, and only a small number of custom-machined parts, practitioners are welcome to modify the design and customize our system to their needs.
PAMY2 can be equipped with any electrically driven end-effector~(such as the custom-fabricated articulated 3D-printed hand on the front page). 
We also open-source an electrically driven two-DoF wrist that can be combined with, for instance, a racket for table tennis~(see~\cref{fig:wristracket}).  

The main contribution of this work is the design of a robot arm 
(i)~that is less prone to damage upon collision due to its lightweight construction, passively compliant actuation, and tendon drives,
(ii)~with enhanced ease of control, achieved by minimizing nonlinearities, primarily caused by high friction,
(iii)~that allows for repeatable dynamic motions facilitating the collection of large amounts of data for long-term training,
(iv)~designed for replicability and adaptability, which allows researchers to build upon and customize our robot for their specific research question.

In \cref{sec:pamy}, we present the key design decisions that accomplish these objectives.
These include a description of the tendon-driven design and the choices that reduce friction in the tendons and joints and increase the system's robustness.
In \cref{sec:experiments}, we conduct experiments to demonstrate our system's effectiveness. 
We perform long-term dynamic motions to verify robustness and conduct measurements to quantify impact safety. 
To showcase enhanced ease of control, we demonstrate the increased linearity of our robot. 
Finally, we apply our system to a challenging dynamic table tennis task, illustrating its capability for rapid yet precise movements.

\section{Related work}
\label{sec:related_work}

\paragraph{Safety through collision avoidance}

Safety in robotics is typically tackled by instrumenting the environment with sensors to detect and track humans and obstacles in the workspace.
Sensors employed for this purpose range from distance sensors mounted to the robot~\cite{avanzini_safety_2014} to depth cameras~\cite{schmidt_depth_2014} and marker-based motion capture systems~\cite{lasota_toward_2014}.
If the distance between the robot and a human or obstacle is below a threshold, a safety controller adapts the robot motion to avoid a collision~\cite{schmidt_depth_2014} or slows down and stops the motion~\cite{lasota_toward_2014,rybski_sensor_2012}.
These methods generally come with additional costs for the sensors and the need for sensor calibration.
Due to the potential for occlusions, they typically require multiple sensors that capture the scene from different angles.
Furthermore, collision avoidance strategies tend to be very conservative because they aim to avoid collisions at all cost, resulting in a robot that is heavily constrained in its motions.

Robot safety is particularly important when training policies via RL.
During training, the RL agent typically explores random actions.
Due to the uncontrolled nature of these exploration strategies, the resulting motions can be dangerous for both the robot and its environment.
Safe RL aims to mitigate these safety concerns by discouraging the agent from visiting unsafe states.
To that end, these methods modify, e.g., the optimization objective~\cite{heger_consideration_1994,borkar_q_2002,geibel_risk_2005,basu_learning_2008}, the exploration behavior~\cite{garcia_safe_2012,geramifard_intelligent_2013,berkenkamp_safe_2017,chow_lyapunov_2018}, or the action selected by the policy~\cite{dalal_safe_2018,pham_optlayer_2018,shao_reachability_2021}.
Unless provided with additional domain knowledge, safe RL methods need to explore dangerous states at least once during training to learn that these states are unsafe~\cite{garcia_comprehensive_2015}.
Domain knowledge, e.g., in the form of a dynamics model or an expert policy, might not always be available, and visiting an unsafe state even once can already cause severe damage to the robot or its environment.

\paragraph{Safety through compliance}
Inherently compliant robots are a viable option to alleviate some safety requirements and avoid the extensive use of sensors for collision avoidance and the dependence on domain knowledge for safe RL.
Soft robot components, like passively compliant joints~\cite{park_safe_joint_2008} or links~\cite{park_safe_link_2008}, can significantly reduce contact forces upon collision.
\citet{gealy_quasi_2019} built a 7-DoF robot arm that achieves passive compliance through backdrivable transmissions.
GummiArm~\cite{stoelen_co-exploring_2016,stoelen_gummiarm_2022} and BioRob~\cite{lens_biorob_2010} 
are notable examples that employ elastic tendons for mechanical compliance.
Both systems leverage tendon-driven architectures to enhance compliance and safety.

GummiArm is characterized by its agonist-antagonist actuation system, employing pairs of opposing elastic tendons that control the movement and stiffness of each joint via variable co-contraction levels. Such a design allows the arm to absorb impacts through its joints' natural flexing, reducing the risk during accidental human contact. The elastic materials used for the tendons contribute both to the safety and the bio-fidelity of the arm, allowing it to execute smooth movements.

BioRob focuses on incorporating a series elastic actuatuation concept within a highly lightweight structure. These actuators introduce significant compliance at each joint, serving both to mitigate impact forces and to enhance energy efficiency through the storage and release of kinetic energy during tasks. BioRob’s design emphasizes minimal moving mass and compact actuator integration.

An alternative is offered by PAMs, which are inherently compliant actuators that can achieve high forces.
These actuators are widely used in human-inspired robot arms~\cite{tondu_seven_2005,boblan_humanoid_2010,ikemoto_direct_2012,ikemoto_humanlike_2012,gong_bionic_2019}.
Passively compliant robots can also be combined with active collision avoidance strategies, such as in \cite{schiavi_integration_2009}.
In such a combination, the robot's compliance enhances safety in the event of undetected or unavoidable collisions.

\begin{figure*}[t]
    \centering
    \scriptsize
        \centering
        \subfloat[Entire arm]{
    \includegraphics[scale=0.18,keepaspectratio,angle=90]{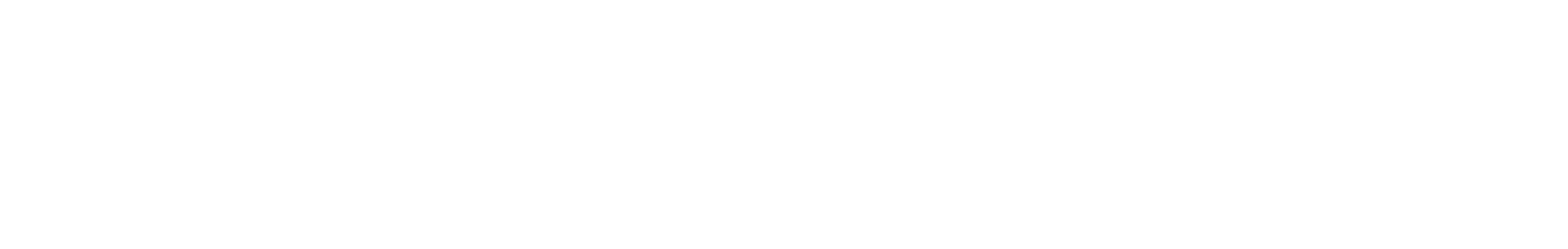}
        \label{sfig:pamy_arm}}
    \hfill
    \begin{minipage}[t]{0.86\textwidth}
        \centering
        \vspace{-15cm}
        \hspace{4.5cm}
        \subfloat[Bowden tubes]{
                    \centering\scriptsize
                                        \includegraphics[scale=0.15]{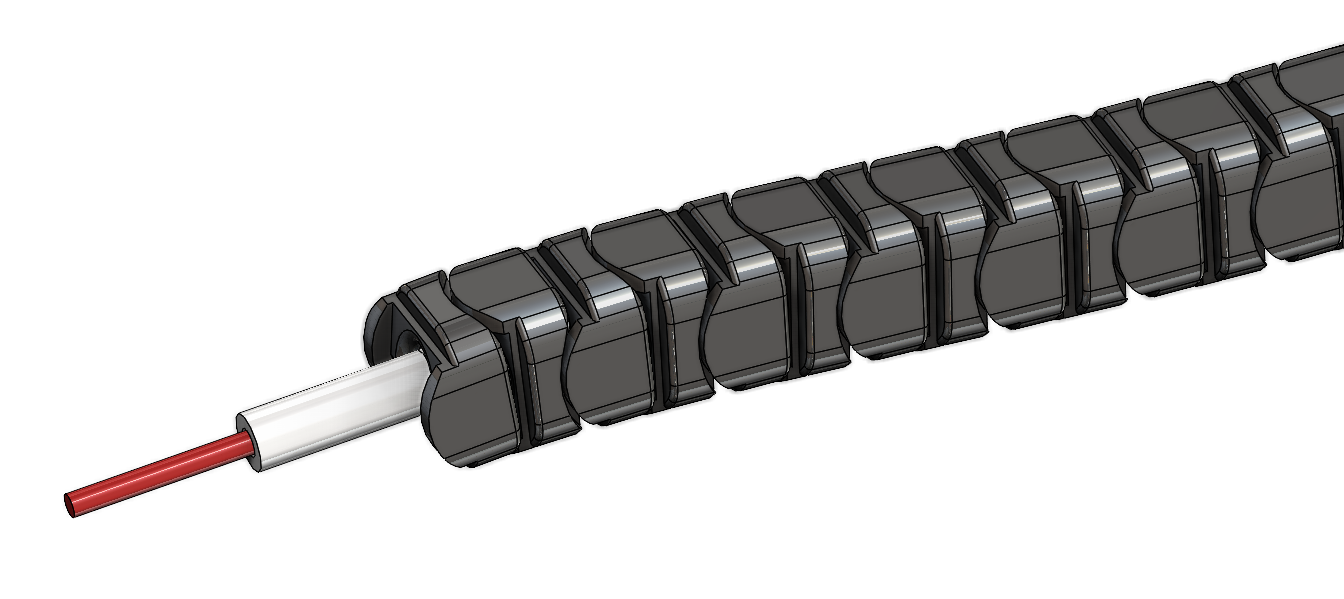}
                    \label{sfig:bowden_2}
                                    }\\[-0.8cm]
        \subfloat[Upper elbow joint]{
            \centering\scriptsize
                        \includegraphics[scale=0.19]{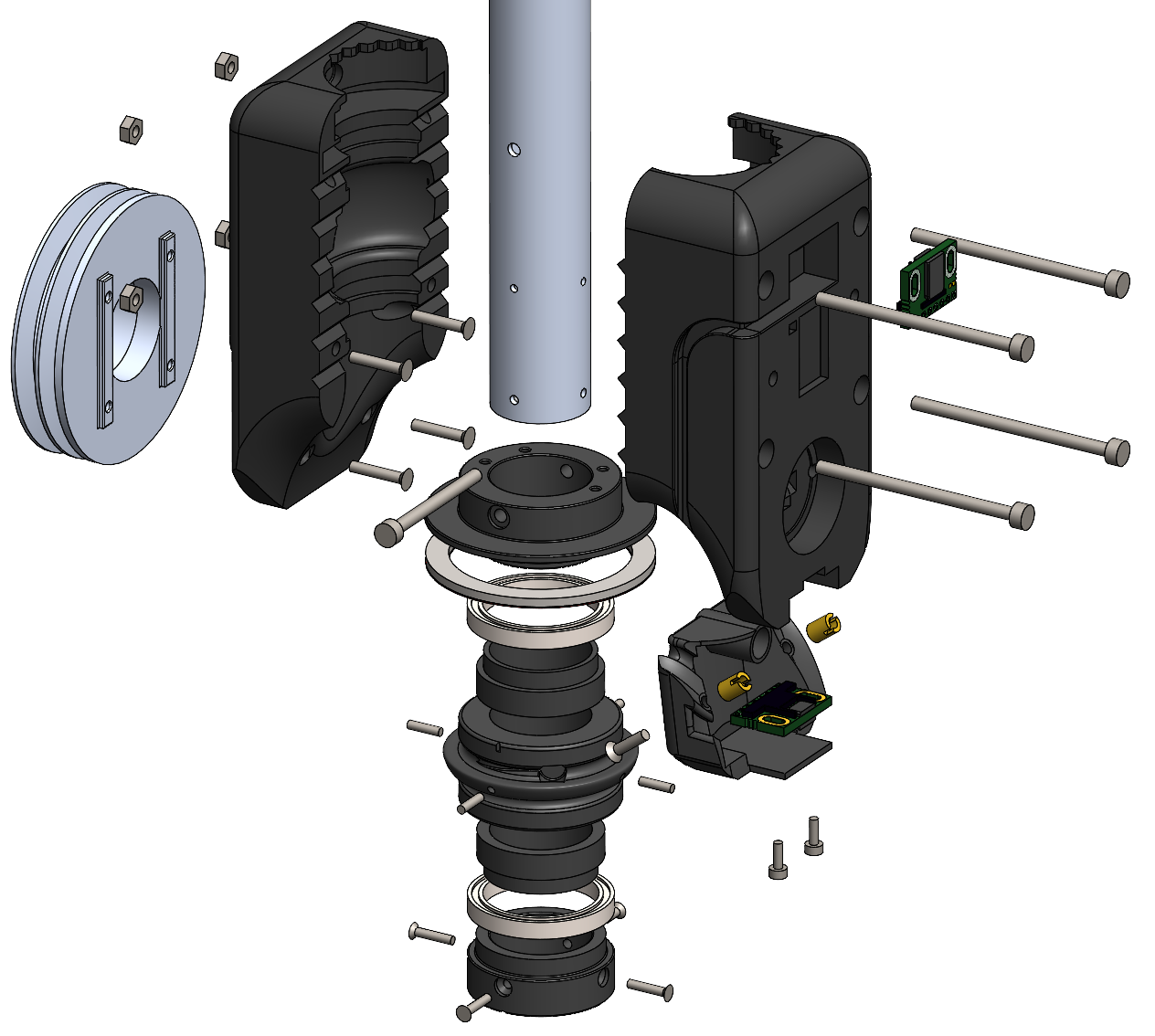}
            \label{sfig:dof12_1}
        }\hfill
        \subfloat[Lower elbow joint]{
            \centering\scriptsize
                        \includegraphics[scale=0.15]{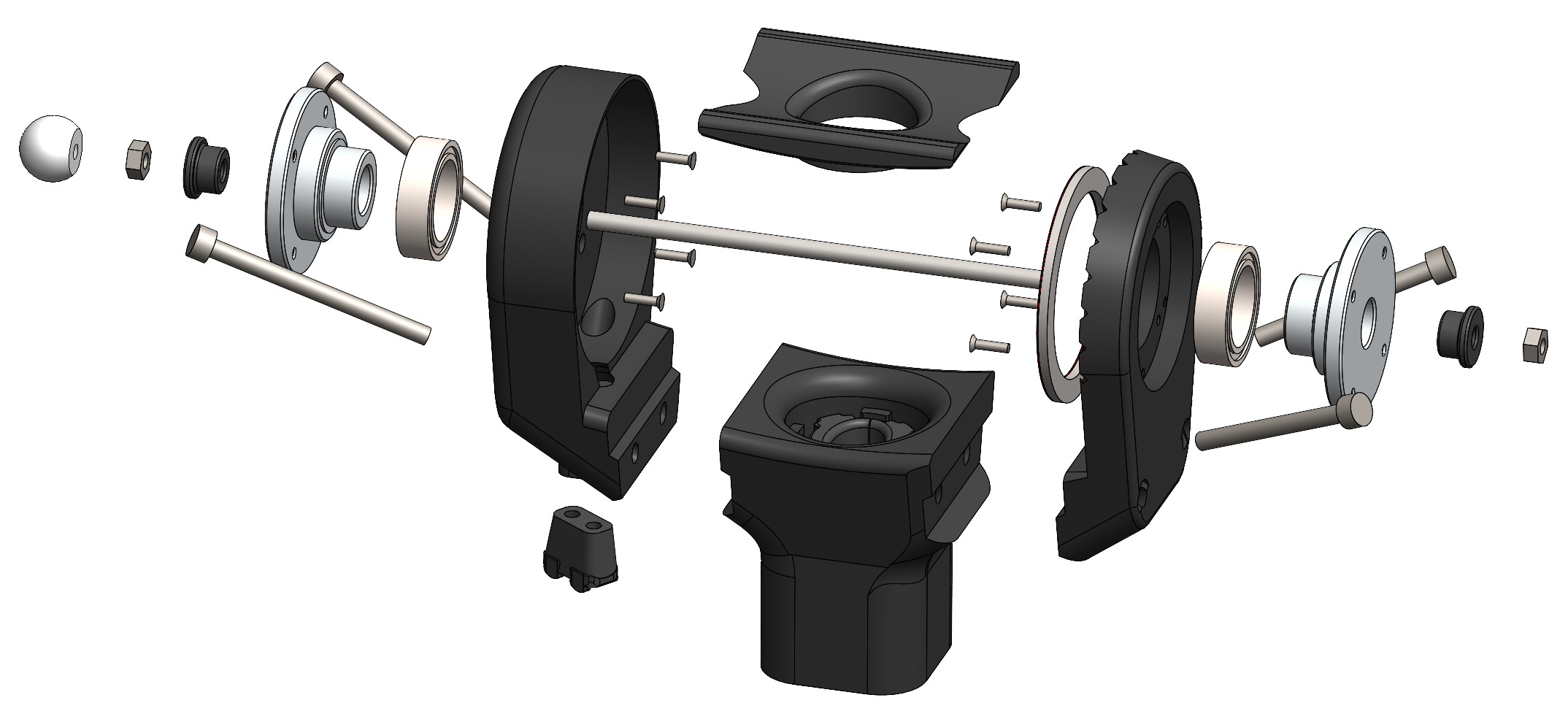}
            \label{sfig:dof12_2}
        }\\[0.32cm]
        \centering
        \hspace{1.5cm}
        \subfloat[Lower base joint]{
            \centering\scriptsize
                        \includegraphics[scale=0.16]{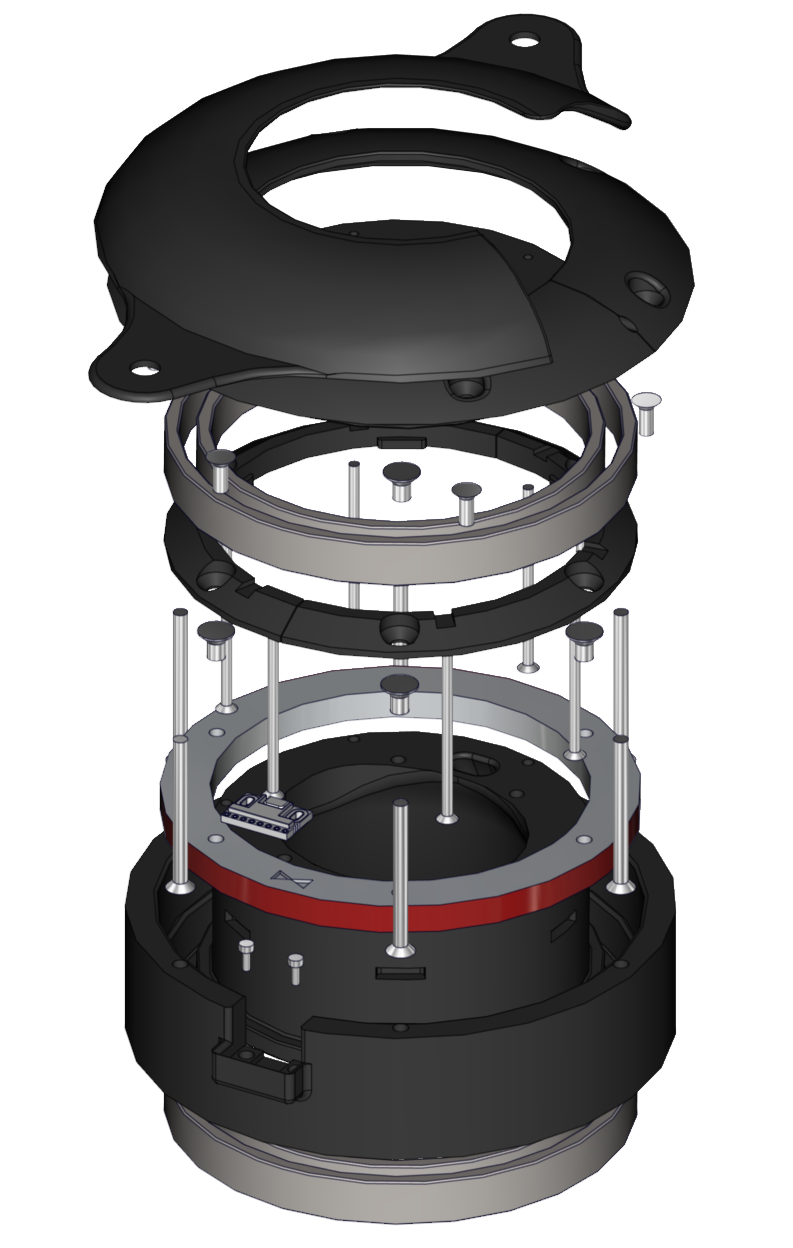}
            \label{sfig:dof34_1}
        }
        \hfill
        \subfloat[Upper base joint]{
            \centering\scriptsize
            \includegraphics[scale=0.21]{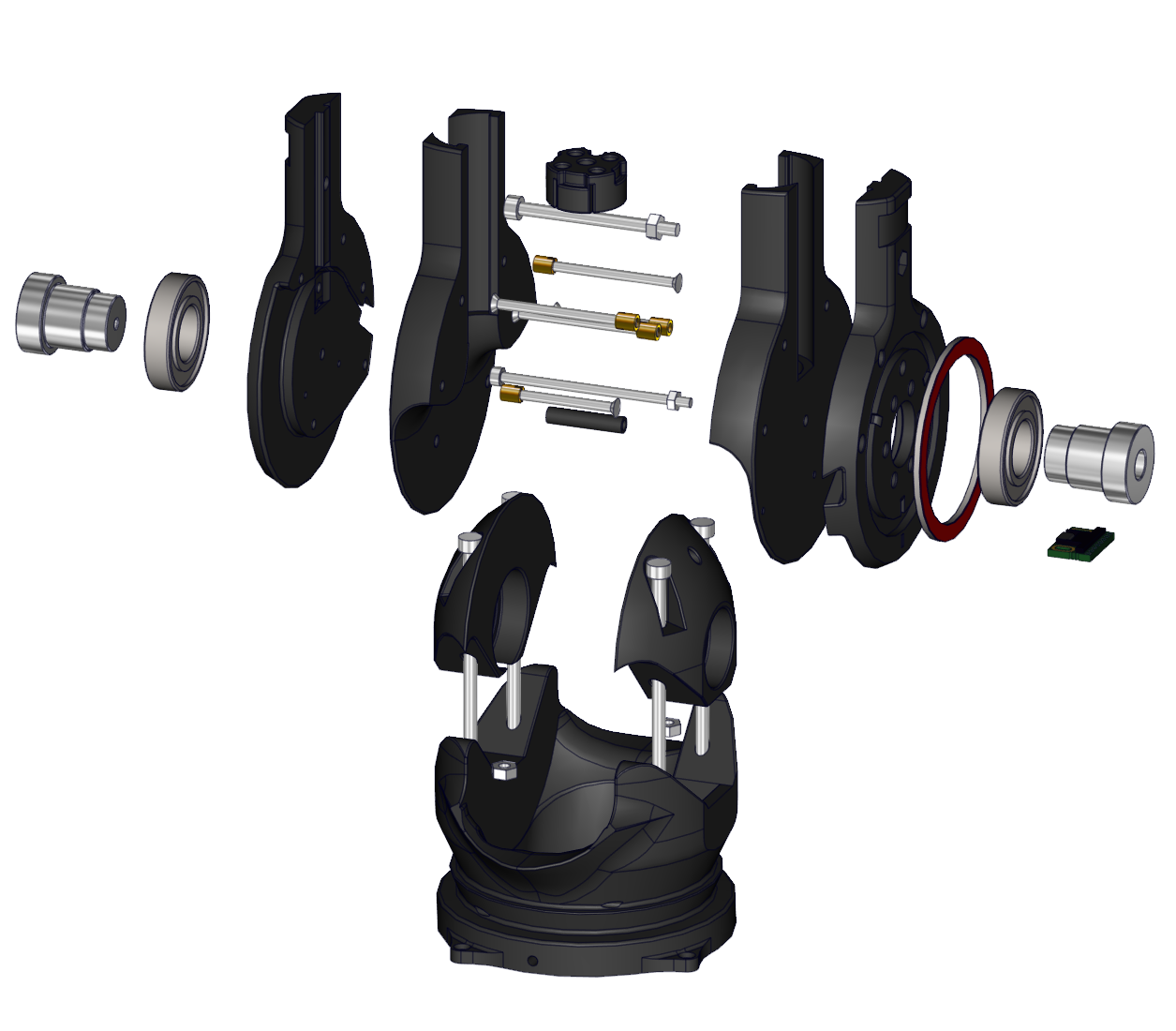}
            \label{sfig:dof34_2}
        }
        \hspace{1cm}

    \end{minipage}

    \vspace{0.2cm}

    \begin{tikzpicture}[remember picture, overlay, blend mode=darken]
        \node (img1) at (-0.25, 7.48) {\includegraphics[scale=0.465]{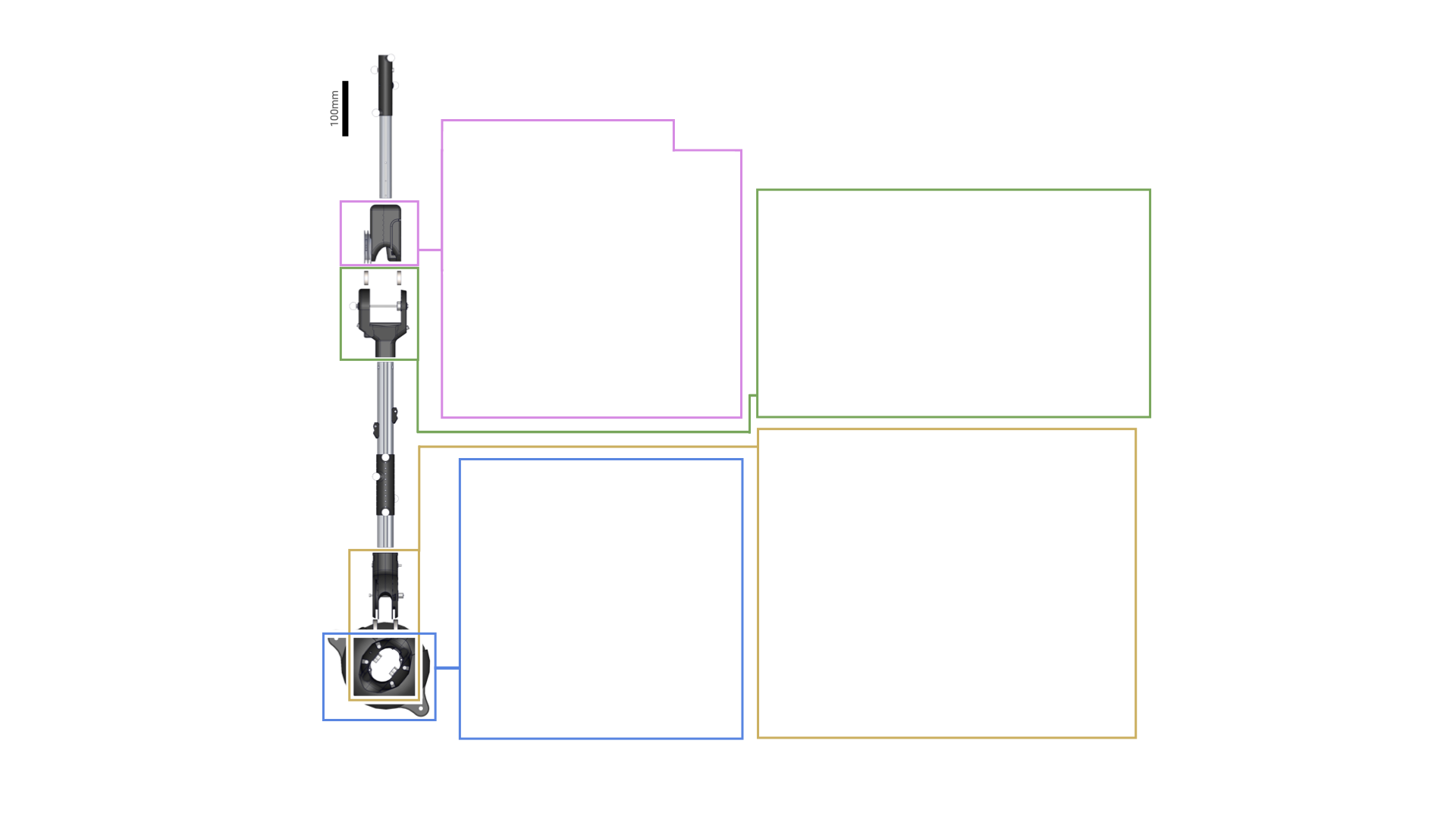}};
    \end{tikzpicture}

    \caption{Design of the tendon-driven robot arm (a). The arm has a rotational and a swivel DoF within the first (e), (f), and second joint (c), (d). It features ball bearings, which are low in friction. Many parts are self-designed and 3D-printed, which are shown colored in black. The four angle encoders are shown with a small green circuit board. The bowden tubes (b) guide the tendons from the muscles to the joints. They feature an inner tube and outer support elements that help maintain constant tendon length.}
    \label{fig:design}
\end{figure*}

\paragraph{Generating dynamic motions}
Safety through compliance is essential in dynamic tasks because collisions are more likely at high velocities, and impacts are more severe due to the high momentum of the robot.
However, fast motions apply additional strain to the system and, therefore, few robot designs in the literature are capable of generating dynamic motions.
\Citet{ikemoto_humanlike_2012} evaluate their design on a dynamic throwing task.
\Citet{mori_high_2018} designed a high-speed robot for a highly dynamic badminton task, which achieves racket speeds of \SI{21}{m/s}.

BioRob~\cite{lens_biorob_2010} and GummiArm~\cite{stoelen_co-exploring_2016} are also both capable of dynamic motions. BioRob arm is able to achieve end effector velocities of \SI{7.4}{m/s}. Analysis of Figure 10 in \cite{stoelen_co-exploring_2016}, showing ballistic movements, suggests GummiArm is slower than BioRob. This compares to PAMY2's end effector velocities of \SI{12}{m/s} during the table tennis experiments shown in Section~\ref{subsec:exp_dyn_task}.
GummiArm also has its actuators distributed along the arm, potentially increasing moving mass and impacting safety at high speeds or necessitating the use of lighter, less powerful actuators.
Furthermore, neither BioRob nor GummiArm extensively focus on reducing internal friction within their systems. This is particularly relevant when considering their dynamic motion capabilities and overall ease of control.
To our knowledge, long-term experimental evaluations, which are critical for assessing durability and reliability, have not been reported for these systems, so it remains unclear whether these motions can be executed robustly over long periods.

\section{Realization of PAMY2}
\label{sec:pamy}

In this section, we present the design of our robot, PAMY2, which substantially improves upon the tendon-driven robot introduced by \citeauthor{buchler_lightweight_2016}~\cite{buchler_lightweight_2016,buchler_control_2018}, referred to as PAMY1 throughout the paper.
We detail the improvements to the mechanical design, Bowden tubes, bearings, and pneumatics.
Furthermore, we discuss our design choices in light of the goals of impact safety, robustness, and ease of control.
The mechanical design of the arm is depicted in \cref{fig:design}.

\subsection{Design Choices to Improve Impact Safety}
Because collisions cannot always be avoided without limiting the performance of the robot, one of the primary design objectives is to ensure that the impact of such collisions is limited. We achieve this goal by incorporating a tendon-driven design and leveraging passively compliant actuators.
As shown in \cref{fig:design}, the actuators of our robot are not on the robot arm but at the base.
Therefore, the moving masses (about \SI{1.3}{kg}) and inertia are small compared to traditional robot designs where actuators are typically located at the joints.

Active and passive compliance are two distinct approaches to achieving compliance. Active compliance utilizes sensor data and feedback control to adapt and respond to external forces.
One example of active compliance are collision reaction schemes.
However, the reaction times are often too long to prevent damage.
In contrast, passive compliance is achieved through the inherent mechanical properties of the robot, such as elastic joints or soft materials.
Our robot achieves passive compliance through the use of PAM actuators.
These actuators are inherently compliant, allowing the robot to absorb and dissipate external forces without the need for complex control schemes. 

\subsection{Design Choices to Increase Ease of Control and Extend Durability}
Reducing friction is essential for improving the robot's ease of control, as it reduces uncertainties and nonlinearities in the dynamics.
Furthermore, friction leads to wear, which limits the system's longevity.
This section highlights the design choices that help minimize friction in our robot.

\subsubsection{Bowden Tubes and Tendons}
Improving the Bowden tubes is key to addressing friction, durability, and maintenance challenges commonly encountered in tendon-driven systems.
Our design incorporates continuous Polytetrafluoroethylene (PTFE) Bowden tubes, which have very low friction. These tubes also exhibit resistance to kinking and separation minimizing the risk of tendon entanglement. To further reduce friction within the PTFE tube, we use Ballistol universal oil to lubricate the tendon strings.
In addition, our system features a design consisting of an inner tube and custom outer support elements, as depicted in \cref{sfig:bowden_2}. These outer support elements were specifically engineered for easy 3D printing and manufactured out of Onyx, a carbon-fiber reinforced polyamide filament that withstands exceptionally high forces without breaking. At the same time, this new design fulfills the usual task of a Bowden tube: ensuring a constant tendon length during arm movements by providing external support. Consequently, the movement of one joint influences others only through the rigid body dynamics rather due to the tendon drives, improving overall ease of control.

The tendons themselves are made from a Dyneema string that has high strength-to-weight ratio and durability. Dyneema strings are chosen due to their maximum load capacity, and smooth surface, which reduces friction inside the guiding tubes. The tendon strings used in our robot have a diameter of \SI{1.8}{mm}, but can withstand a load capacity of \SI{500}{daN}. To improve Dyneema's creep resistance under high load, we use a pre-tensioned version of the string that is heat-treated to reduce elongation to less than $1\%$, enhancing its load capacity compared to similar strings of the same diameter and ensuring minimal length change under tension. The lower temperature resistance of Dyneema is mitigated by our design focusing on reducing friction. The effectiveness of this design in reducing heat is shown in experiment
\ref{subsubsec:friction}.
For the tendon connections, we utilize a knot-based method, which circumvents the disadvantages of adhesive-based attachments, such as long drying times and potential weakening over time. To address the decreased tear-strength of strings because of the tightening of the knot, our knotting method involves guiding the tendon along a rounded curve before knotting.
More details about the parts and materials used can also be found on our project website.

\subsubsection{Bearings}
The previous design of \citeauthor{buchler_lightweight_2016}~\cite{buchler_lightweight_2016,buchler_control_2018} utilizes gliding bearings, which offer the benefit of being highly compact.
However, these bearings exhibit considerable friction and stiction.
In contrast, PAMY2 includes industry-standard ball bearings at the shoulder and elbow joints to significantly reduce the friction and stiction, while also providing increased off-axis rigidity and improved longevity.  The primary tradeoff is a slight increase in the mass of the joints and packaging complexity.

\subsection{Improved Pneumatics}
In the pneumatic system of our tendon-driven robot arm, we have implemented several optimizations to enhance performance and reliability.
First, we employed optimized tube routing to improve airflow and reduce pressure losses, particularly by avoiding sharp 90-degree angles in the pneumatic lines.
Second, we incorporated a buffer reservoir to stabilize the air pressure supply in front of the valves, ensuring consistent and efficient actuation. Lastly, we designed a ring circuit for the pneumatic system, further improving the air pressure distribution among the valves.

\subsection{Open-Source Hardware and Software}
To facilitate further research and development in dynamic robotic tasks, we have made both the hardware and software components of our robot open-source.
\subsubsection{Hardware}
Our approach aims at enabling others to build upon our work and adapt the robot for their specific applications.
Therefore, the design primarily employs off-the-shelf, commercially available components, 3D-printed parts, and only a limited number of custom-machined parts, thus making it cheaper than many other industrial or research robots.
The total material costs amount to approximately €14,185, broken down as follows: Base at €2,652, Arm at €1,448, Electronics at €4,053, and Pneumatics at €6,032.
It is important to note, however, that the 3D-printed components require specialized printers, 
capable of reinforcing parts with continuous fibers.

\begin{wrapfigure}{r}{0.2\textwidth} 
\vspace{-20pt}
  \begin{center}
    \includegraphics[width=0.19\textwidth]{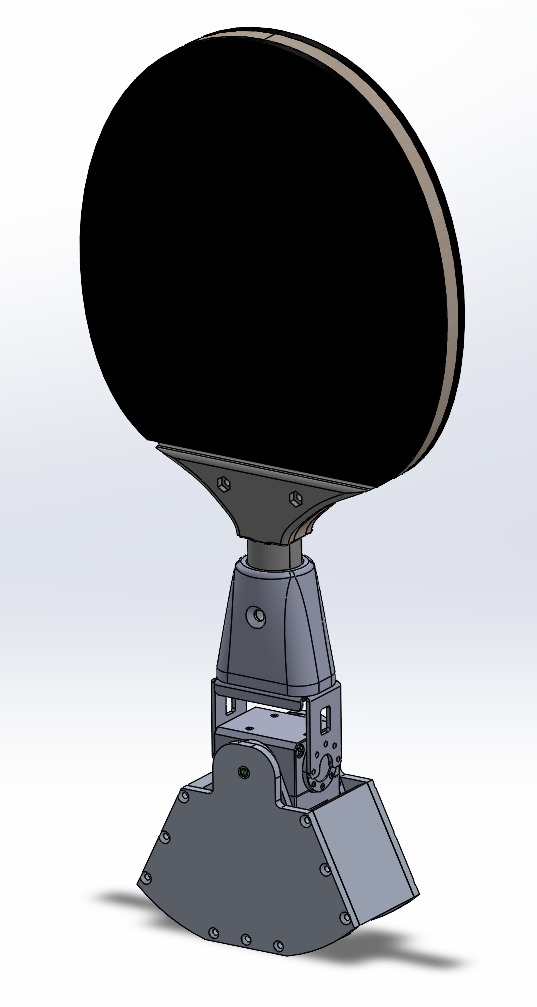}
    \caption{Electrically driven wrist for PAMY2 that can be combined with an end-effector}
    \label{fig:wristracket}
  \end{center}
  \vspace{-20pt}
  \end{wrapfigure} 

\subsubsection{End-effectors}
PAMY2 can be combined with any electrically driven end-effector, since we feed cables seamlessly through the inside of the arm. 
In this manner, hands could be mounted, such as the custom-fabricated articulated 3D-printed hand on the front page.
We also open-source a two-DoF wrist that features position and torque control and can be combined with, for instance, a racket for table tennis~\cref{fig:wristracket}.

\subsubsection{Software}
We provide an open-source software framework with a versatile API in Python and C++ for controlling and monitoring the robot, based on the o80 framework~\cite{berenz2021o80}. The o80 software framework interfaces with the robot's Programmable Logic Controller~(PLC). Communication between the PLC and the o80 software running on the PC is facilitated through UDP, transmitting data such as the robot state (joint angles and velocities, muscle pressures, and valve positions), actions (target pressures or target joint positions, depending on the control mode), and error information.

\begin{figure}[t]
    \centering\scriptsize
    \includegraphics[scale=0.08]{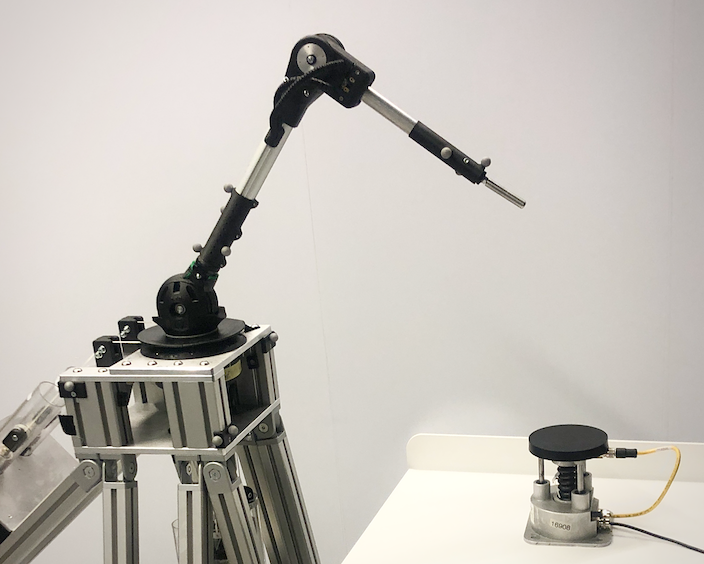}
    \vspace{0.1cm}
    \caption{Experimental setup for the collision force measurements. A Pilz PRMS is mounted to a table onto which the end effector of the robot is colliding.}
    \label{fig:prms_setup}
\end{figure}

\section{Experiments \& Evaluations}
\label{sec:experiments}

\begin{figure}[t]
    \centering\scriptsize

    \subfloat[PAMY2]{
        \centering\scriptsize
        \begin{tikzpicture}

\begin{axis}[
height=9.57968316831683cm,
tick align=outside,
tick pos=left,
width=4.4cm,
x grid style={white},
xlabel={contact condition},
xmajorgrids,
xmin=0.5, xmax=10.5,
xtick style={color=black},
xtick={1,2,3,4,5,6,7,8,9,10},
ytick={0.12, 0.26, 0.4, 0.54, 0.68, 0.82, 0.96, 1.1, 1.24, 1.38, 1.52, 1.66, 1.8, 1.94},
axis line style={draw=none},
xticklabel style={ yshift=0.14cm},
yticklabel style={ xshift=0.14cm},
tick style={draw=none},
y grid style={white},
ylabel={contact velocity [m/s]},
ymajorgrids,
ymin=0.05, ymax=1.975,
ytick style={color=black}
]
\addplot graphics [includegraphics cmd=\pgfimage,xmin=0.5, xmax=10.5, ymin=0.05, ymax=1.975] {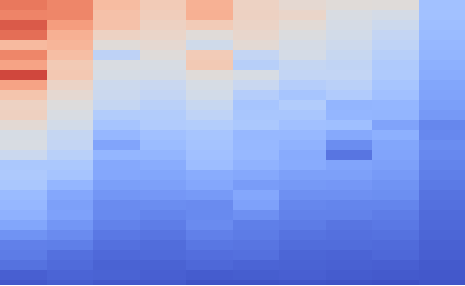};
\end{axis}

\end{tikzpicture}
        \label{sfig:impact_pamy2}
    }
    \begin{minipage}{0.6\textwidth}
            \vspace{-7.78cm}
        \subfloat[Panda]{
        \centering\scriptsize
        \begin{tikzpicture}

\begin{axis}[
colorbar horizontal,
colormap={mymap}{[1pt]
  rgb(0pt)=(0.2298057,0.298717966,0.753683153);
  rgb(1pt)=(0.26623388,0.353094838,0.801466763);
  rgb(2pt)=(0.30386891,0.406535296,0.84495867);
  rgb(3pt)=(0.342804478,0.458757618,0.883725899);
  rgb(4pt)=(0.38301334,0.50941904,0.917387822);
  rgb(5pt)=(0.424369608,0.558148092,0.945619588);
  rgb(6pt)=(0.46666708,0.604562568,0.968154911);
  rgb(7pt)=(0.509635204,0.648280772,0.98478814);
  rgb(8pt)=(0.552953156,0.688929332,0.995375608);
  rgb(9pt)=(0.596262162,0.726149107,0.999836203);
  rgb(10pt)=(0.639176211,0.759599947,0.998151185);
  rgb(11pt)=(0.681291281,0.788964712,0.990363227);
  rgb(12pt)=(0.722193294,0.813952739,0.976574709);
  rgb(13pt)=(0.761464949,0.834302879,0.956945269);
  rgb(14pt)=(0.798691636,0.849786142,0.931688648);
  rgb(15pt)=(0.833466556,0.860207984,0.901068838);
  rgb(16pt)=(0.865395197,0.86541021,0.865395561);
  rgb(17pt)=(0.897787179,0.848937047,0.820880546);
  rgb(18pt)=(0.924127593,0.827384882,0.774508472);
  rgb(19pt)=(0.944468518,0.800927443,0.726736146);
  rgb(20pt)=(0.958852946,0.769767752,0.678007945);
  rgb(21pt)=(0.96732803,0.734132809,0.628751763);
  rgb(22pt)=(0.969954137,0.694266682,0.579375448);
  rgb(23pt)=(0.966811177,0.650421156,0.530263762);
  rgb(24pt)=(0.958003065,0.602842431,0.481775914);
  rgb(25pt)=(0.943660866,0.551750968,0.434243684);
  rgb(26pt)=(0.923944917,0.49730856,0.387970225);
  rgb(27pt)=(0.89904617,0.439559467,0.343229596);
  rgb(28pt)=(0.869186849,0.378313092,0.300267182);
  rgb(29pt)=(0.834620542,0.312874446,0.259301199);
  rgb(30pt)=(0.795631745,0.24128379,0.220525627);
  rgb(31pt)=(0.752534934,0.157246067,0.184115123);
  rgb(32pt)=(0.705673158,0.01555616,0.150232812)
},
height=3.708cm,
point meta max=350,
point meta min=0,
tick align=outside,
tick pos=left,
width=4.4cm,
x grid style={white},
xlabel={contact condition},
xmajorgrids,
xmin=0.5, xmax=10.5,
xtick style={color=black},
xtick={1,2,3,4,5,6,7,8,9,10},
ytick={0.12, 0.26, 0.4, 0.54},
axis line style={draw=none},
xticklabel style={ yshift=0.14cm},
yticklabel style={ xshift=0.14cm},
tick style={draw=none},
y grid style={white},
ylabel={contact velocity [m/s]},
ymajorgrids,
ymin=0.05, ymax=0.575,
ytick style={color=black}
]
\addplot graphics [includegraphics cmd=\pgfimage,xmin=0.5, xmax=10.5, ymin=0.05, ymax=0.575] {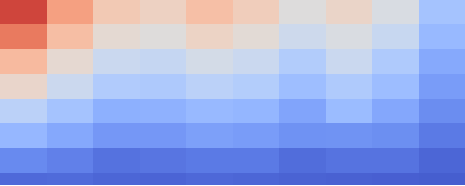};
\pgfplotsset{every colorbar/.append style={
at=(current axis.north),
anchor=south,
yshift=1.0cm,
width=1.0*\pgfkeysvalueof{/pgfplots/parent axis width},
height=0.25cm,
align=center,
colormap name=mymap,
axis line style={draw=none},
colorbar style={
every inner x axis line/.style={draw=none}, 
xtick={0,100,200,300},
xticklabels={0,100,200,350},
xticklabel style={font=\tiny},
scaled ticks=false,
ticks=none,
draw=white,
},
}}
\end{axis}
\node[above=0.95cm of current axis.north, text=black] {peak impact force [N]};

\end{tikzpicture}
        \label{sfig:impact_panda}
    }\\[0.8cm]
    \subfloat[UR5e]{
        \centering\scriptsize
        \begin{tikzpicture}

\begin{axis}[
height=3.708cm,
tick align=outside,
tick pos=left,
width=4.4cm,
x grid style={white},
xlabel={contact condition},
xmajorgrids,
xmin=0.5, xmax=10.5,
xtick style={color=black},
xtick={1,2,3,4,5,6,7,8,9,10},
ytick={0.12, 0.26, 0.4, 0.54},
axis line style={draw=none},
xticklabel style={ yshift=0.14cm},
yticklabel style={ xshift=0.14cm},
tick style={draw=none},
y grid style={white},
ylabel={contact velocity [m/s]},
ymajorgrids,
ymin=0.05, ymax=0.575,
ytick style={color=black}
]
\addplot graphics [includegraphics cmd=\pgfimage,xmin=0.5, xmax=10.5, ymin=0.05, ymax=0.575] {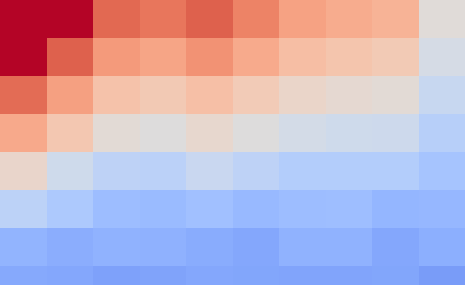};
\end{axis}

\end{tikzpicture}
        \label{sfig:impact_ur5e}
    }
    \end{minipage}
    \vspace{0.1cm}
    \caption{Collision force map depicting peak impact forces resulting from varying impact velocities and contact scenarios for our robot, alongside the Franka Emika Panda and the Universal Robot UR5e for comparison. Our findings reveal that our robot, when operating at high velocities, generates impact forces akin to those exhibited by the other two robots at considerably lower velocities.
    We express our gratitude to \citet{kirschner2021towards} for generously sharing the impact data for the Panda and UR5e robot arms.}
\label{fig:force_map}
\end{figure}

\setlength{\figwidth }{0.9\columnwidth} 
\setlength{\figheight }{.309\figwidth}
\begin{figure}[t]
	\centering
	\vspace{-.25cm}
        \scriptsize
	\subfloat[Experimental setup]{
		\centering
		\includegraphics[scale=0.065, trim={0 0 5cm 0},clip]{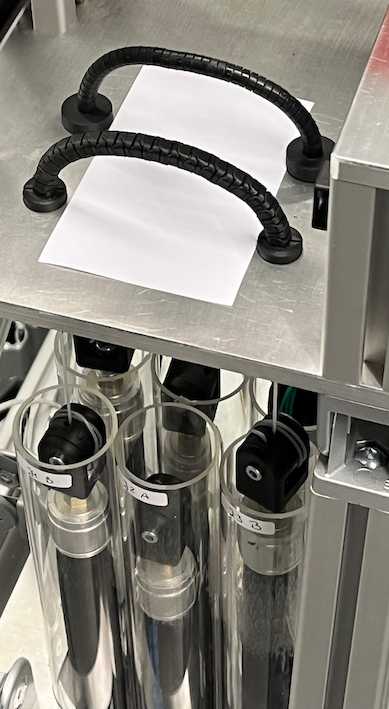}
		\label{sfig:bowden_experiment_setup}
	}
	\begin{minipage}[t]{0.6\textwidth}
            \vspace{-6.5cm}
		\subfloat[Thermal image]{
			\centering\scriptsize
			\scalebox{1.0}{\input{figures/thermal-img}}
			\label{sfig:bowden_experiment_results_image}
		}\\[0.29cm]
		\subfloat[Temperature measured over time]{
			\centering\scriptsize
			\scalebox{1.0}{\begin{tikzpicture}

\definecolor{darkgray176}{RGB}{176,176,176}
\definecolor{lightgray204}{RGB}{204,204,204}
\definecolor{olivedrab10516779}{RGB}{105,167,79}
\definecolor{royalblue73133231}{RGB}{73,133,231}

\begin{axis}[
height=3.4cm,
legend cell align={left},
legend columns=2,
legend style={
  fill opacity=0.8,
  draw opacity=1,
  text opacity=1,
  at={(0.5,1.35)},
  anchor=north,
  draw=lightgray204,
  /tikz/every odd column/.append style={column sep=0.05cm},
  /tikz/every even column/.append style={column sep=0.25cm}
},
mark size=1.5pt,
tick align=outside,
tick pos=left,
width=5.5cm,
x grid style={darkgray176},
xlabel={time [min]},
xmin=-1.5325, xmax=32.1825,
xtick style={color=black},
y grid style={darkgray176},
ylabel={temperature [\SI{}{\celsius}]},
ymin=23, ymax=58,
ytick style={color=black}
]
\addplot [draw=olivedrab10516779, fill=olivedrab10516779, mark=*, only marks]
table{
x  y
12.25 54.49
16.35 54.28
22.4833333333333 55.17
3.05 42.91
28.6166666666667 55.4
30.65 55.53
6.11666666666667 50.24
14.3 55.15
19.4166666666667 54.79
18.3833333333333 54.44
2.03333333333333 38.42
20.4333333333333 55.09
7.15 51.4
27.5833333333333 55.63
26.5666666666667 55.77
8.16666666666667 52.37
10.2166666666667 53.73
15.3166666666667 55.09
24.5166666666667 55.29
9.18333333333333 53.09
5.1 48.67
11.2333333333333 53.65
29.6333333333333 55.92
17.3666666666667 54.79
4.08333333333333 46.37
25.55 55.8200000000001
1.01666666666667 31.14
0 25.87
21.45 55.09
13.2833333333333 54.69
23.5 55.25
};
\addlegendentry{PAMY1}
\addplot [draw=royalblue73133231, fill=royalblue73133231, mark=*, only marks]
table{
x  y
12.25 39.02
16.35 39.65
22.4833333333333 39.13
3.05 33.39
28.6166666666667 39.78
30.65 40.23
6.11666666666667 37.17
14.3 39.89
19.4166666666667 39.26
18.3833333333333 39.22
2.03333333333333 31.56
20.4333333333333 39.56
7.15 37.58
27.5833333333333 40
26.5666666666667 40.02
8.16666666666667 38.21
10.2166666666667 38.81
15.3166666666667 39.74
24.5166666666667 39.8
9.18333333333333 38.45
5.1 36.42
11.2333333333333 39.0700000000001
29.6333333333333 40.3200000000001
17.3666666666667 39.47
4.08333333333333 34.89
25.55 40.08
1.01666666666667 28.39
0 25.42
21.45 39.35
13.2833333333333 39.15
23.5 39.84
};
\addlegendentry{PAMY2}
\end{axis}

\end{tikzpicture}}
			\label{sfig:bowden_experiment_results_graph}
		}
	\end{minipage}
    \vspace{0.15cm}
    \caption{Experiment comparing the friction of our Bowden tubes (top) with those utilized by~\cite{buchler_lightweight_2016} (bottom).
The experimental setup is illustrated in (a): Both types of Bowden tubes are actuated by rapidly switching muscles of the same type.
The thermal camera's heat map (b) and the temperature evolution over time (c) both show a significantly lower temperature increase for our Bowden tubes.
As the temperature increase is caused by friction, this finding implies that our Bowden tubes exhibit significantly lower friction. 
}
	\label{fig:bowden_experiment}
\end{figure}

\begin{figure*}[t]
\centering
\input{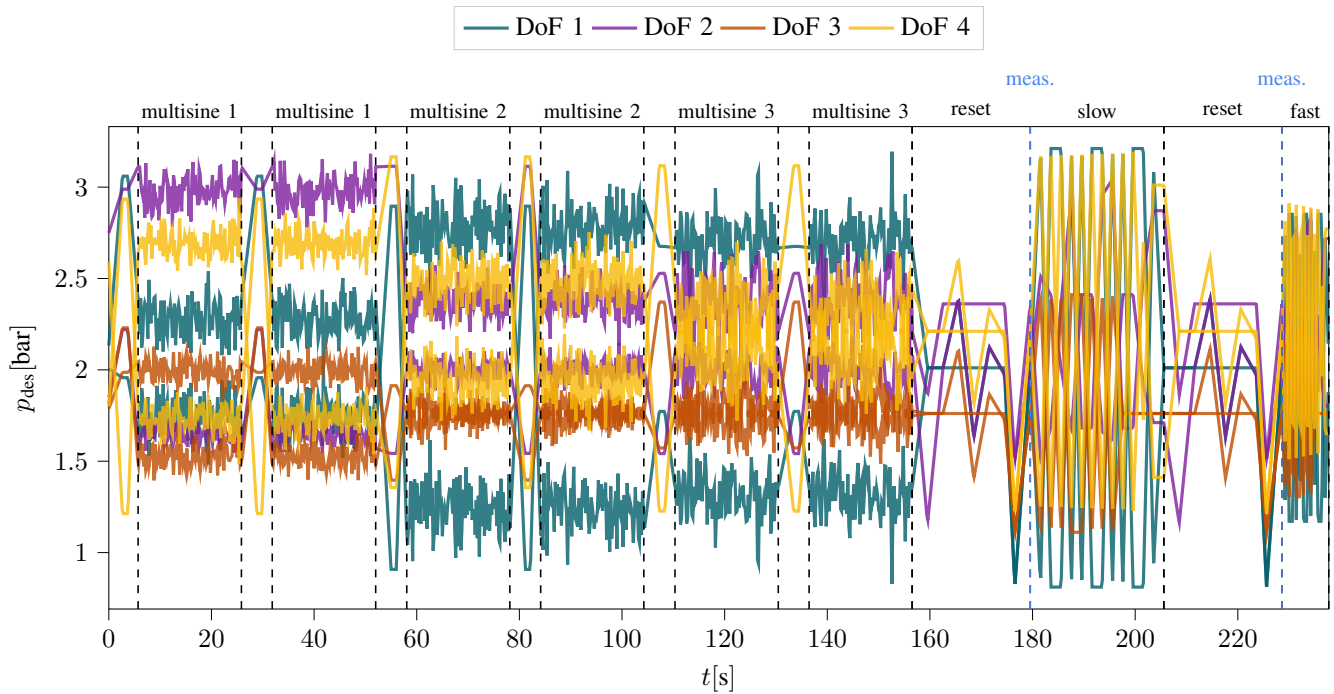}
\caption{Target pressures during an episode of the long-term experiment. All actions are executed open-loop. An episode consists of multisine signals with a frequency of up to 10Hz which are randomly sampled before each episode to explore different areas of the state space. A reset motion sequence is aimed at minimizing the robot's final position dependence on its previous state. At the end of this reset sequence, the repeatability measurement is taken to assess PAMY2's consistency and repeatability across the duration of the experiment. Finally, there are movements executed from sets of fixed target pressures at lower and higher speed.}
\label{fig:long_term_1ep}
\end{figure*}

\begin{figure}[tb]
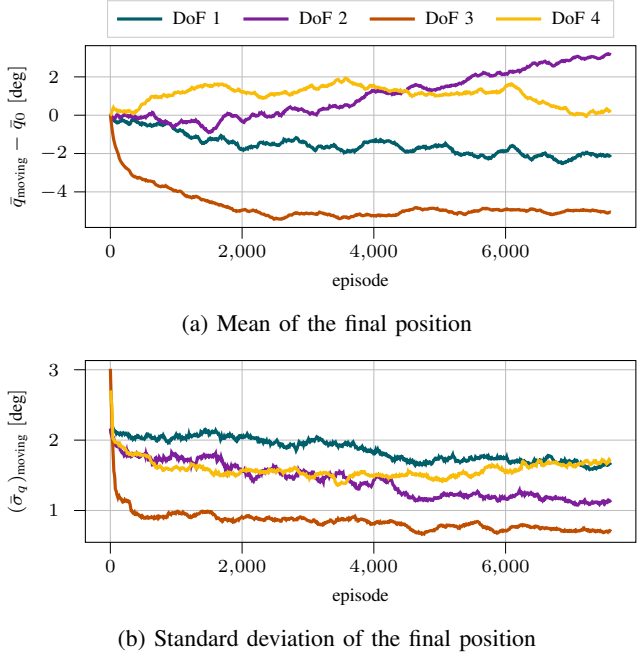

\centering
\vspace{-.25cm}
\subfloat[Mean of the final position]{
    \centering\scriptsize
    \hspace{-.25cm}
    \scalebox{1.0}{\input{figures/plot_long_term_mean_s600}}
    \label{sfig:long_term_mean}
}\\
\vspace{0.25cm}
\subfloat[Standard deviation of the final position]{
    \centering\scriptsize
    \hspace{-.25cm}
    \scalebox{1.0}{\input{figures/plot_long_term_error_s600}}
    \label{sfig:long_term_error}
}
\vspace{0.15cm}
\caption{
Moving mean position relative to the initial mean position $\bar{q}_{\textrm{moving}} - \bar{q}_0$ and the standard deviation $(\bar{\sigma}_q)_{\textrm{moving}}$ of the robot's final position after the open-loop reset motion during the approximately 25 days of the long-term experiment. It uses a moving average window of 400 episodes. A stable mean position (a) alongside a small and slightly decreasing standard deviation (b) over time indicates consistent performance throughout the long-term experiment.
}
\label{fig:long_term}
\end{figure}

In this section, we present a series of experiments designed to assess the efficacy of our new robot arm. Our experiments focus on evaluating the following characteristics: Impact safety, robustness, ease of control, and the ability to perform rapid and precise movements.

\subsection{Evaluating Impact Safety}

\begin{table}[b]
\centering
\caption{Definition of contact conditions based on ISO/TS 15066~\cite{ISO15066}.}
\begin{tabular}{ccccc}
\toprule
\textbf{No.} & \textbf{Body part} & \textbf{Stiffness} & \textbf{Hardness} & \textbf{Pain Threshold} \\ \midrule
1   & Skull       & 150 N/mm  & 70 ShA   & 130 N     \\
2   & Face/hand   & 75 N/mm   & 70 ShA   & 65 N      \\
3   & Lower legs  & 60 N/mm   & 30 ShA   & 260 N     \\
4   & Thighs      & 50 N/mm   & 30 ShA   & 300 N     \\
5   & Neck        & 50 N/mm   & 70 ShA   & 440 N     \\
6   & Lower arms  & 40 N/mm   & 70 ShA   & 320 N     \\
7   & Back        & 35 N/mm   & 30 ShA   & 420 N     \\
8   & Upper arms  & 30 N/mm   & 30 ShA   & 300 N     \\
9   & Chest       & 25 N/mm   & 70 ShA   & 280 N     \\
10  & Abdomen     & 10 N/mm   & 10 ShA   & 220 N     \\ \bottomrule
\end{tabular}
\label{table:stiffness}
\end{table}

One of the main goals of our robot arm is to ensure superior impact safety compared to traditional motor-driven systems. We achieve this objective primarily through the tendon-driven design, which relocates the heavy actuators to the robot base.
Although the use of compliant actuators contributes to the overall safety of the robot, the system's inertia primarily determines the peak force at impact.

To evaluate the impact safety of our robot, we examine the peak force occurring during potential collisions. For our experiments, we employ the Pilz Robot Measurement System (PRMS) to measure forces during collisions. This device comprises a one-dimensional load cell, a spring, and a rubber cover. Various springs and covers are available to adjust the stiffness and hardness according to different human body parts as specified by ISO/TS 15066:2016~\cite{ISO15066}. This technical specification introduces a model of the human body, covering 21 body regions. For each body region, it provides contact conditions and a pain tolerance, which we display in \cref{table:stiffness}. 

We compare our measurements to the results obtained in previous studies by \citet{kirschner2021towards, kirschner_experimental_2021}.
To ensure that our measurements accurately reflect the impact safety of our robot, we measure the generated forces at a position close to the robot center.
This position ensures that most of the robot's mass contributes to the peak force experienced during a collision. \Cref{fig:prms_setup} illustrates the robot's position during the measurements.
To collect the data, we execute trajectories with linear changes in muscle pressure.
Upon detecting a sudden change in velocity, indicating a collision, we halt the movement by keeping the target pressure fixed.
By modifying the rate of change of the target pressure, we evaluate the impact safety of our robot at various velocities.

\Cref{fig:force_map} shows peak forces during collisions for our tendon-driven arm and for the Franka Emika Panda and Universal Robots UR5e, two conventional motor-driven systems, investigated in prior research. The results clearly show the superior impact safety of our robot, as it achieves similar contact forces while moving at almost four times the speed.

However, it is important to emphasize that although our tendon-driven arm is significantly safer compared to other robotic arms, it could still cause serious injuries if a human is struck with full force at high speeds.
Future work should focus on further enhancing the safety of our system by combining its inherently safer hardware design with algorithmic approaches, additional sensors, or safety mechanisms.

\subsection{Evaluating Robustness}
\label{subsec:verifying_robustness}

\begin{figure*}
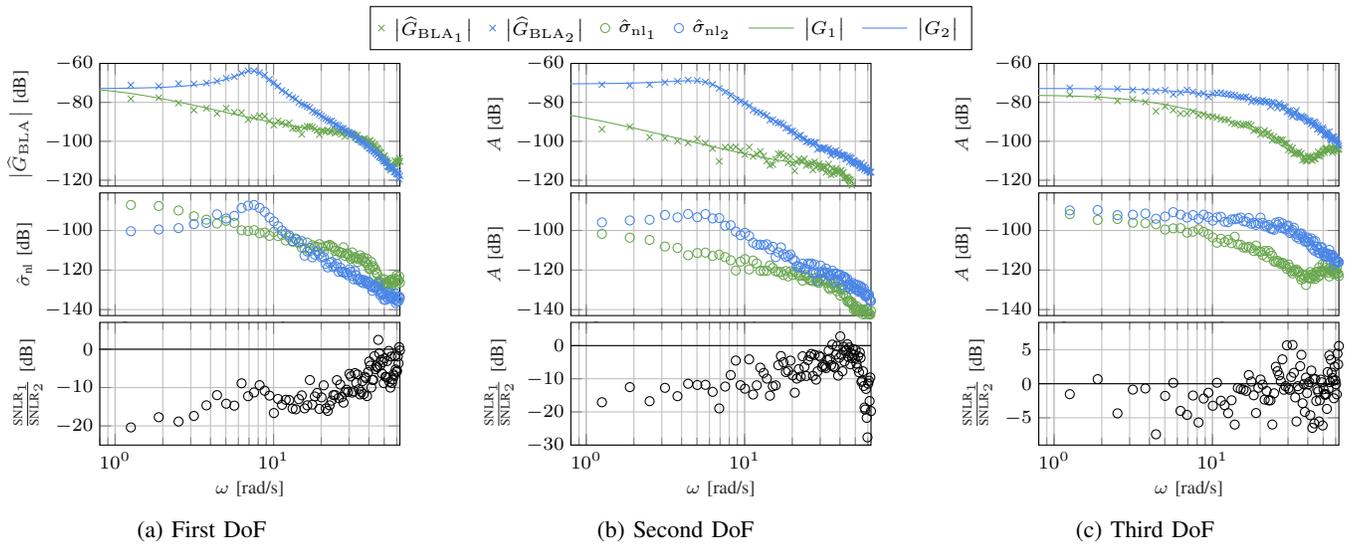

        \begin{minipage}{\textwidth}
        \centering
        \definecolor{mycolor1}{RGB}{106,168,80}
\definecolor{mycolor2}{RGB}{74,134,232}
\definecolor{mycolor3}{RGB}{106,168,80}
\definecolor{mycolor4}{RGB}{74,134,232}
\definecolor{goldenrod21018910}{RGB}{106,168,80}

\begin{tikzpicture}
\begin{axis}[
    hide axis,
    width=0.20\textwidth,
    ymin=-50,
    ymax=10,
    xmin=-50,
    xmax=10,
    legend columns=6,
    legend style={
        at={(0,0)},anchor=north, 
        /tikz/every even column/.append style={column sep=0.15cm},
        /tikz/every odd column/.append style={column sep=0.05cm}
    },
    font=\scriptsize,
    mark size=1.7pt
]
\addlegendimage{only marks, mark=x, mark options={solid, draw=mycolor1}}
\addlegendentry{$\big| \widehat{G}_{\rm{BLA_1}} \big|$}
\addlegendimage{only marks, mark=x, mark options={solid, draw=mycolor2}}
\addlegendentry{$\big| \widehat{G}_{\rm{BLA_2}}\big|$}
\addlegendimage{only marks, mark=o, mark options={solid, draw=mycolor3}}
\addlegendentry{$\hat{\sigma}_{\rm{nl_1}}$}
\addlegendimage{only marks, mark=o, mark options={solid, draw=mycolor4}}
\addlegendentry{$\hat{\sigma}_{\rm{nl_2}}$}
\addlegendimage{goldenrod21018910}
\addlegendentry{$\big| G_{\rm{1}} \big|$}
\addlegendimage{mycolor2}
\addlegendentry{$\big| G_{\rm{2}} \big|$}
\end{axis}
\end{tikzpicture}
        \vspace{.25cm}
        \begin{minipage}{\textwidth}
    	\subfloat[First DoF]{
    		\centering\scriptsize%
                \input{figures/res_1_new}
    		\label{sfig:linearity_results_ampl}
    	}
            \hfill
    	\subfloat[Second DoF]{
    		\centering\scriptsize%
                    \input{figures/res_2_new}
    		\label{sfig:linearity_results_nonlin}
    	}
           \hfill
    	\subfloat[Third DoF]{
    		\centering\scriptsize%
                    \input{figures/res_3_new}
    		\label{sfig:linearity_results_snlr}
    	}
        \end{minipage}
        \end{minipage}
	\caption{
        Comparison of the linearity between PAMY1 (index 1) and PAMY2 (index 2). The new system demonstrates increased amplitude ($A$) and a higher bandwidth across all DoFs (first row). The second row shows the absolute level of nonlinearity of the two systems. To make this nonlinearity between the systems comparable, we examine the SNLR (third row), which is significantly lower for the first two DoFs at most frequencies, and marginally lower for the third DoF at lower frequencies. These findings underscore the enhanced tracking performance and reduced nonlinearity of the new system.
	}
	\label{fig:linearity_results}
\end{figure*}

Reinforcement learning enables robots to achieve high performance in complex tasks. However, for these approaches, training for long durations is crucial. 
We aim to produce a system that lasts and is reliable and, therefore, minimize the main contributor to failure: friction. 
\subsubsection{Friction Quantification}
\label{subsubsec:friction}

Friction plays a significant role in tendon-driven robots as it converts kinetic energy into thermal energy.
Because it is widely acknowledged that using ball bearings instead of sliding contact bearings is an effective way to reduce friction, we focus our experiments on evaluating the other major source of friction in our tendon-driven robot: the Bowden tubes.

Higher friction implies that more heat is generated in the Bowden tubes.
Therefore, we compare the friction in our Bowden tubes with the original system from \cite{buchler_lightweight_2016}  by capturing the heat generated during operation with a thermal camera.
We set up an experiment where two muscles are connected directly using a Bowden tube, as shown in \cref{sfig:bowden_experiment_setup}. The antagonistic muscle pair is contracted in an alternating manner at a frequency of \SI{3.33}{\Hz} for 30 minutes.
\Cref{sfig:bowden_experiment_results_image} displays the thermal image captured at the end of the experiment, which indicates that our Bowden tube remained substantially cooler than the original.
A more detailed analysis of the temperature change in the two Bowden tubes can be seen in \cref{sfig:bowden_experiment_results_graph}. The data points in the plot represent the median temperature of the five hottest pixels in the corresponding Bowden tube's image region. The graph illustrates that the temperature of the original tubes rises more rapidly and ultimately converges to a higher value. These measurements confirm that the design of our Bowden tubes offers significantly lower friction than the original system.

\subsubsection{Long-term Dynamic Motions}

Learning complex skills with real robots often necessitates numerous interactions with the environment. To showcase our robot's capability in collecting the data required for mastering dynamic tasks, we conducted a long-term experiment involving various dynamic movements. This experiment evaluates the system's reliability and robustness over extended periods.

We designed an array of movement patterns, encompassing random multisine signals, fixed target pressure movements with varying time intervals, and reset motions. The reset motion, executed in an open-loop fashion without feedback from joint position measurements, involves moving to medium pressures, then to minimum pressures, and finally to medium pressures again. This particular reset motion was chosen to make the final position more independent of the preceding position or motion, thus making differences in this position more indicative of changes in hardware.
\Cref{fig:long_term_1ep} displays the actuation signal for one episode of the long-term experiment.

During the experiment, the robot operated continuously for approximately 25 days, amassing a comprehensive dataset of diverse robotic motions. The data was recorded at a high sampling rate of \SI{500}{Hz}. The dataset includes the observed and desired pressure for each muscle, the position, and velocity of each joint, as well as timestamps.

Throughout the long-term experiments, the robot exhibited a high level of robustness and reliability, with no significant signs of wear or damage. This outcome underscores the effectiveness of our design in ensuring the robot's durability during dynamic tasks.

To quantitatively assess the system's repeatability, we analyzed the robot's final position after executing the reset motion multiple times. The small deviation in the final positions, illustrated in \cref{fig:long_term}, indicates that the system maintains consistent performance even after prolonged usage. This repeatability is crucial for successful reinforcement learning, as it ensures the robot can reliably execute learned policies.

\subsection{Evaluating Ease of Control}

\subsubsection{Increased Linearity}

We employ the method proposed by~\citet{ma_learning-based_2022} for system identification in the frequency domain to quantify the nonlinearity in PAMY1 and PAMY2.
Each degree of freedom is treated as a single-input and single-output (SISO) system.
We design ten different excitation signals with the same frequency spectrum and ten randomly chosen phase spectra, exciting the frequency lines $\Omega = \{0.1 \text{Hz},0.2 \text{Hz},\dots,10 \text{Hz}\}$.

We excite each degree of freedom individually. Each excitation signal is applied for ten periods continuously, and the response signals of the first two periods are discarded to avoid the effect of transients.
Let $U^{i}\mleft(j \omega_k\mright)$ and $Y^{i}\mleft(j \omega_k\mright)$ with $i=1,\dots,p$ and $\omega_k \in \Omega$ denote the discrete Fourier transformation (DFT) of the input (difference in target pressure for antagonistic muscle pairs) and output signals (joint angles) of the $i$-th period, respectively, where $p$ represents the total number of periods after discarding (here $p=8$), and $j=\sqrt{-1}$ denotes the imaginary number.
First of all, we calculate the average value of the input and output signals in the frequency domain over all the different periods.
\begin{align}
    \hat{Y}\mleft(j \omega_k\mright) &= \frac{1}{p} \sum_{i=1}^{p} Y^i\mleft(j \omega_k\mright) \\
    \hat{U}\mleft(j \omega_k\mright) &= \frac{1}{p} \sum_{i=1}^{p} U^i\mleft(j \omega_k\mright)
\end{align}

\noindent The average frequency response function (FRF) $\widehat{G}$ is given by
\begin{align}
    \widehat{G}\mleft(j \omega_k\mright) &= \frac{\widehat{Y}\mleft(j \omega_k\mright)}{\widehat{U}\mleft(j \omega_k\mright)}. \label{eq:frf}
        \end{align}

Since the identified system is nonlinear, the discrepancy between the measured average FRFs that arises when having excitation signals with different phase spectra provides a means to characterize the nonlinearities. If the system were linear, then applying excitation signals with different phase realizations would not affect the average FRF.
First, we calculate the corresponding average FRF $\widehat{G}^{l},l=1,\dots,m$
for each input signal according to \cref{eq:frf}, where the superscript $l$ refers to the different excitation signals (here $m=10$). Then, the average FRF over all excitation signals is given by
\begin{equation}
\widehat{G}_{\text{BLA}}\mleft(j \omega_k\mright) = \frac{1}{m} \sum_{l=1}^{m}\widehat{G}^{l} \mleft(j \omega_k\mright), 
\end{equation}
where the subscript BLA refers to ``Best Linear Approximation''. Lastly, an estimate of the system's nonlinearity is given by
\begin{equation}
    \hat{\sigma}^2_{\text{nl}} \mleft(j \omega_k\mright) = \frac{1}{m(m-1)} \sum_{l=1}^{m} \left|\widehat{G}^{l}\mleft(j \omega_k\mright) - \widehat{G}_{\text{BLA}} \mleft(j \omega_k\mright)\right|^2,
\end{equation}
where the subscript nl refers to ``nonlinearity''.

We note that the absolute value of $\hat{\sigma}^2_{\text{nl}}$ is not a good indicator since it would be affected by a simple re-scaling of the output variable.
To better compare the degree of nonlinearity of the two systems, we define the signal-to-nonlinearity ratio~(SNLR) as
\begin{equation}
    \textrm{SNLR} \mleft(j \omega_k\mright) = \left(\frac{\left|\widehat{G}_{\text{BLA}}\mleft(j \omega_k\mright)\right|}{\hat{\sigma}_{\text{nl}}}\right)^2.
\end{equation}

\Cref{fig:linearity_results} displays the system identification results.
We observe that the new system exhibits higher amplitudes compared to the old system, indicating improved tracking of dynamic motions.
At the same time, each system's degree of nonlinearity and amplitude show the same trend.
The third row shows the ratio of the SNLR for the two systems.
We notice that in a large portion of the frequency spectrum, the SNLR value of the old system is significantly smaller than that of the new system, especially for the first and second DoF.
When adjusted for amplitude, this result shows that the new system has significantly lower nonlinearity than the old system. This indicates that the new system is easier to control, which we will demonstrate in the next experiment.

\subsubsection{Learning a Dynamic Task}
\label{subsec:exp_dyn_task}

\setlength{\figwidth }{0.9\columnwidth} 
\setlength{\figheight }{.309\figwidth}
\begin{figure*}[t]
	\centering
	\vspace{-.25cm}
	\subfloat[Initial position]{
		\centering\scriptsize
		\includegraphics[scale=0.115]{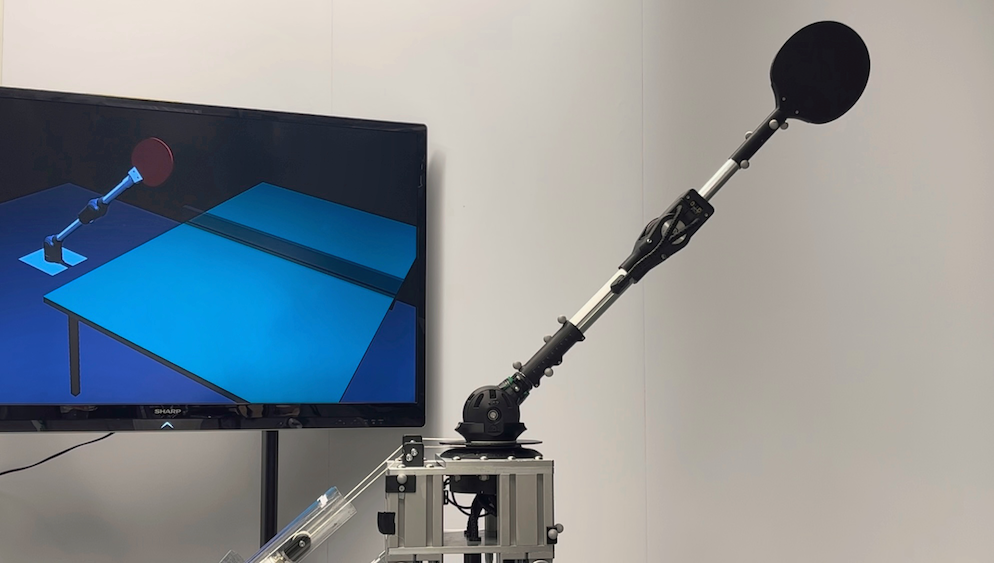}
		\label{sfig:hysr1}
	}\hspace{0.25cm}
        \subfloat[Smashing motion]{
            \centering\scriptsize
            \includegraphics[scale=0.12]{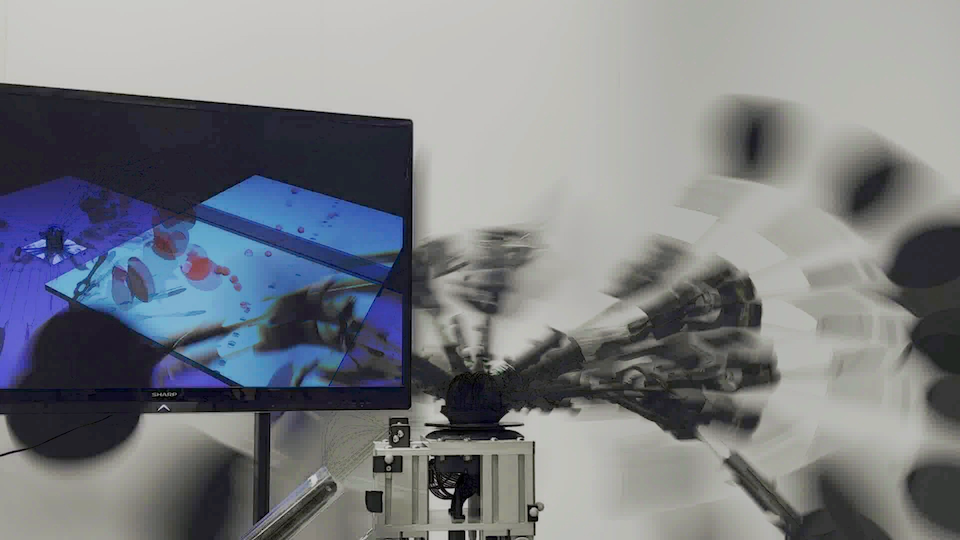}
            \label{sfig:hysr2}
        }
    \vspace{0.15cm}
    \caption{Table tennis smashing experiment with PAMY2. In the HySR setup, we learn with a real robot and a simulated ball. The robot's initial position at the beginning of an episode is shown in (a). During the training, the robot learns a motion in which it first draws back to generate momentum before striking the ball (b). The racket reaches speeds of up to \SI{12}{m/s} during this motion.}
	\label{fig:hysr}
\end{figure*}

\begin{figure*}[t]
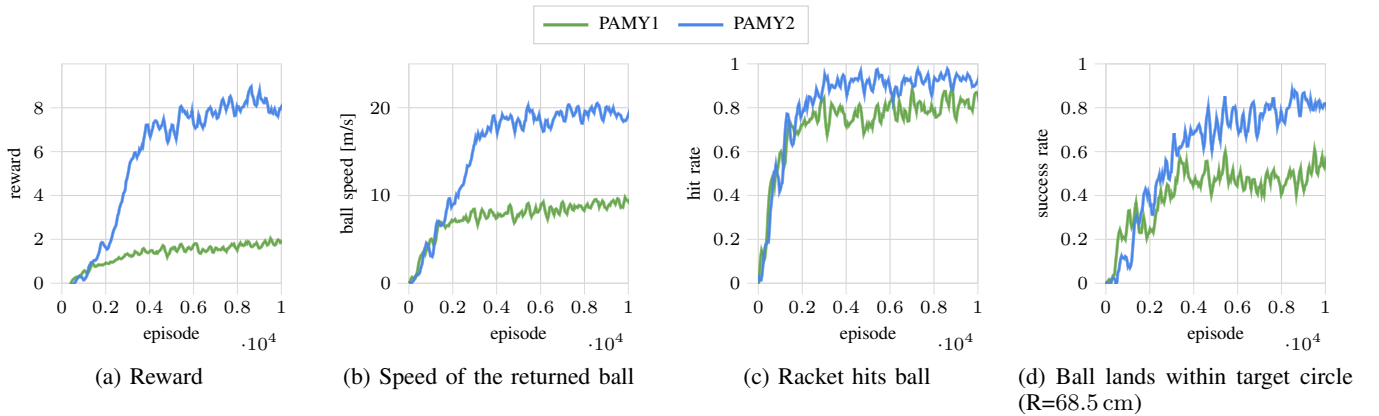

\centering
\definecolor{dodgerblue1101252}{RGB}{74,134,232}
\definecolor{gainsboro}{RGB}{220,220,220}
\definecolor{goldenrod21018910}{RGB}{106,168,80}
\definecolor{lightgray}{RGB}{211,211,211}
\definecolor{lightgray204}{RGB}{204,204,204}

\begin{tikzpicture}
\begin{axis}[
    hide axis,
    width=0.20\textwidth,
    ymin=-50,
    ymax=10,
    xmin=-50,
    xmax=10,
    legend cell align={left},
    legend columns=4,
    legend style={
        fill opacity=0.8,
        draw opacity=1,
        text opacity=1,
        at={(0.5,1.18)},
        anchor=north,
        draw=lightgray204,
        /tikz/every even column/.append style={column sep=0.15cm},
        /tikz/every odd column/.append style={column sep=0.05cm}
    },
    font=\scriptsize,
]
\addlegendimage{goldenrod21018910, very thick}
\addlegendentry{PAMY1}
\addlegendimage{dodgerblue1101252, very thick}
\addlegendentry{PAMY2}
\end{axis}
\end{tikzpicture} \\[0.1cm]
\subfloat[Reward]{
\centering\scriptsize
\scalebox{1.0}{\input{figures/plot_tt_smash_rew}}
\label{sfig:smash_reward}
}
\hfill
\subfloat[Speed of the returned ball]{
\centering\scriptsize
\scalebox{1.0}{\input{figures/plot_tt_smash_vel}}
\label{sfig:smash_velocity}
}
\hfill
\subfloat[Racket hits ball]{
\centering\scriptsize
\scalebox{1.0}{\input{figures/plot_tt_smash_racket_hitting}}
\label{sfig:smash_success_hit}
}
\hfill
\subfloat[Ball lands within target circle (R=\SI{68.5}{cm})]{
\centering\scriptsize
\scalebox{1.0}{\input{figures/plot_tt_smash_table_hitting}}
\label{sfig:smash_success_target}
}
\caption{Results of the table tennis smashing experiment. Compared to the design of \citeauthor{buchler_lightweight_2016}~\cite{buchler_lightweight_2016,buchler_control_2018}, PAMY2 reaches a significantly higher final reward (a). Analyzing this result in detail, we find that it returns balls with higher average speed (b), but at the same time also hits the ball more often (c), and more often the ball lands close to the target position (d). This experiment demonstrates the benefits of the new design for motions that are highly dynamic but also precise.}
\label{fig:smash2}
\end{figure*}

To demonstrate our system's capabilities in a highly dynamic task, we repeat the table tennis smashing experiment from~\cite{buchler_learning_2022}. This task employs a reward function, detailed in \cite{buchler_learning_2022}, that includes both the ball's speed and accuracy concerning a target location on the other side of the table. It is a good demonstration of the robot's potential, as it requires rapid and precise movements, as well as high forces to accelerate the ball to high velocities. Similar to \citet{buchler_learning_2022}, we train the robot in a Hybrid-Sim-and-Real setup~(HySR), with the real robot and a simulated ball as shown in \cref{fig:hysr}. After training, the robot can return real balls with high speeds, albeit training only with simulated balls.

We use a stochastic policy, with actions being changes in target pressures, and use Proximal Policy Optimization (PPO)~\cite{schulman_proximal_2017} as the backbone RL algorithm. Although there were changes in the software, we aimed to keep the setup as similar as possible to~\cite{buchler_learning_2022} to allow for a direct performance comparison. It is worth noting that the learning hyperparameters were optimized for the old system, and we decided not to adapt or optimize them further for our new system to ensure that any performance gains are caused by hardware improvements, not by hyperparameter changes.

As a result of the increased activation bandwidth and also due to the robot discovering an effective strategy of moving along the joint limit of the second DoF before striking the ball, we found that the learning algorithm frequently pushed the robot towards its joint limits, which is detrimental to the robot's longevity. To address this issue and minimize wear and tear on the robot, we introduced a minor term to the reward function that discourages reaching the joint limits. Crucially, aside from this adjustment, no further safety measures are necessary, underscoring the inherent robustness of the PAMY2 design. 

\Cref{fig:smash2} shows that the new design, PAMY2, achieves significantly higher ball speeds than the design of \citeauthor{buchler_lightweight_2016}~\cite{buchler_lightweight_2016,buchler_control_2018}.
Despite the higher ball speeds, it is also more precise in terms of more frequent ball contacts and lower distance to the ball's target location.
When evaluating the trained policy with real balls instead of simulated balls, we achieve similar ball speeds of \SI{20}{m/s}. To our knowledge, this is the fastest robot table tennis play to date and comparable to professional human players~\cite{lee2019speed}.
Overall, PAMY2 reaches performance far superior to PAMY1, demonstrating the benefits of the improved design for highly dynamic and precise motions.

\section{Discussion and Conclusion}
In this paper, we have presented a novel 4-DoF tendon-driven robot arm actuated by PAMs. Our design focuses on reducing friction, passive compliance, and inherent impact safety, allowing the robot to operate efficiently and safely during dynamic tasks.
Through various experiments, we have demonstrated the effectiveness of our robotic arm in terms of these design goals.

Our work contributes to the growing field of soft robotics, which aims to create more adaptable and safer robots, particularly for human-robot interaction scenarios. 
Although our robot arm showcases several advantages over traditional motor-driven systems, there are limitations to our design. Despite the improvements in terms of ease of control, PAM-driven systems still face challenges in achieving the repeatability and precision offered by their motor-driven counterparts. Identifying the optimal set of tasks for our robot arm, where the benefits of safety and dynamic performance outweigh the limitations, is an important avenue for future research.
Recent advances in machine learning for robotics hold great opportunities for enhancing the capabilities of robots like ours. In the table tennis task, we showed that with our improved design, and by leveraging data-driven approaches, it is possible to develop advanced control strategies that address the inherent challenges of PAM-driven systems.

Our work represents a step forward in the development of robotic systems that can achieve high performance while maintaining safety in shared human environments. By making our design and resources open-source, we hope to inspire future research efforts that build upon and refine our work, fostering a new generation of collaborative and versatile robots.

\bibliographystyle{plainnat}
\bibliography{refs}

\end{document}